\documentclass[letterpaper, 10 pt, conference]{ieeeconf}  

\IEEEoverridecommandlockouts                              

\usepackage[ruled,vlined]{algorithm2e}
\usepackage{graphics} 
\usepackage{epsfig} 
\usepackage{gensymb}
\usepackage{cite}
\usepackage{bbding}
\usepackage{makecell}
\usepackage[table]{xcolor}
\usepackage{amsmath} 
\usepackage{multirow}
\usepackage{amssymb}  
\usepackage{xcolor}
\usepackage[pagebackref=true,breaklinks=true,colorlinks,bookmarks=false]{hyperref}

\newcommand{\eg}{\textit{e}.\textit{g}.}

\title{\LARGE \bf
Simultaneous Semantic and Collision Learning for \\6-DoF Grasp Pose Estimation
}

\author{
Yiming Li$^{1,2}$, Tao Kong$^{3}$, Ruihang Chu$^{4}$, Yifeng Li$^{3}$, Peng Wang$^{1,2}$ and Lei Li$^{3}$
\thanks{$^{1}$ Institute of Automation, Chinese Academy of Sciences, Beijing, China.} 
\thanks{$^{2}$ School of Artificial Intelligence, University of Chinese Academy of Sciences, Beijing, China.}
\thanks{$^{3}$ Bytedance AI Lab, Beijing, China.}
\thanks{$^{4}$ The Chinese University of Hong Kong}
\thanks{
Correspondence to: Tao Kong $<$kongtao@bytedance.com$>$ and Peng Wang $<$peng\_wang@ia.ac.cn$>$.}%
}

\begin{document}
\maketitle
\thispagestyle{empty}
\pagestyle{empty}

\begin{abstract}

Grasping in cluttered scenes has always been a great challenge for robots,
due to the requirement of the ability to well understand the scene and object information.
Previous works usually assume that the geometry information of the objects is available, 
or utilize a step-wise, multi-stage strategy to predict the feasible 6-DoF grasp poses.
In this work, we propose to formalize the 6-DoF grasp pose estimation as a simultaneous multi-task learning problem. 
In a unified framework, 
we jointly predict the feasible 6-DoF grasp poses, 
instance semantic segmentation, and collision information. 
The whole framework is jointly optimized and end-to-end differentiable.
Our model is evaluated on large-scale benchmarks as well as the real robot system. 
On the public dataset, 
our method outperforms prior state-of-the-art methods by a large margin (+4.08 AP). 
We also demonstrate the implementation of our model on a real robotic platform and show that the robot can accurately grasp target objects in cluttered scenarios with a high success rate. Project link:  \href{https://openbyterobotics.github.io/sscl}{https://openbyterobotics.github.io/sscl}.

\end{abstract}

\section{Introduction}
It has been a common ability for humans to grasp daily objects. 
By just taking a glance at the scene, we can easily localize the objects of interest and give a proper pose to grasp them. 
However, it remains quite challenging for robots.
The grasping ability requires robots to comprehensively understand the scene and object information.
One of the most challenging parts is being able to estimate the robust,  accurate, and collision-free grasp pose given by visual sensors.

Grasping is a fundamental skill for most robotic manipulation systems,
and has been widely studied over the last decades. 
Previous works on grasp pose estimation could be categorized into two groups: model-based methods and data-driven methods.
Model-based methods assume the geometry information of an object is always available, 
and use mechanical analysis tools~\cite{miller2004graspit,bohg2013data,dang2012semantic} to calculate the grasp poses of an object.
However, it is still an open problem of how to grasp objects with various shapes and sizes in complex scenes.
Data-driven methods that aim to address the generic grasp problem are attaching more and more research attention~\cite{depierre2018jacquard,chu2018real,kumra2017robotic}. 
They usually adopt Deep Neural Networks (DNNs) for the prediction of feasible grasp poses. 
A simple manner is to project the 3D space into a 2D plane, 
transferring the 6-DoF grasping task to a 2D oriented rectangle detection problem~\cite{lenz2015deep,guo2016object}, where the gripper is forced to approach objects from above vertically. 
Although this top-down grasp representation can solve regular grasp tasks, \textcolor{black}{it is still difficult to handle complex objects which are supposed to be grasped from diverse poses in cluttered scenes.}

\begin{figure}[t]
    \centering
    \includegraphics[width=0.5\textwidth]{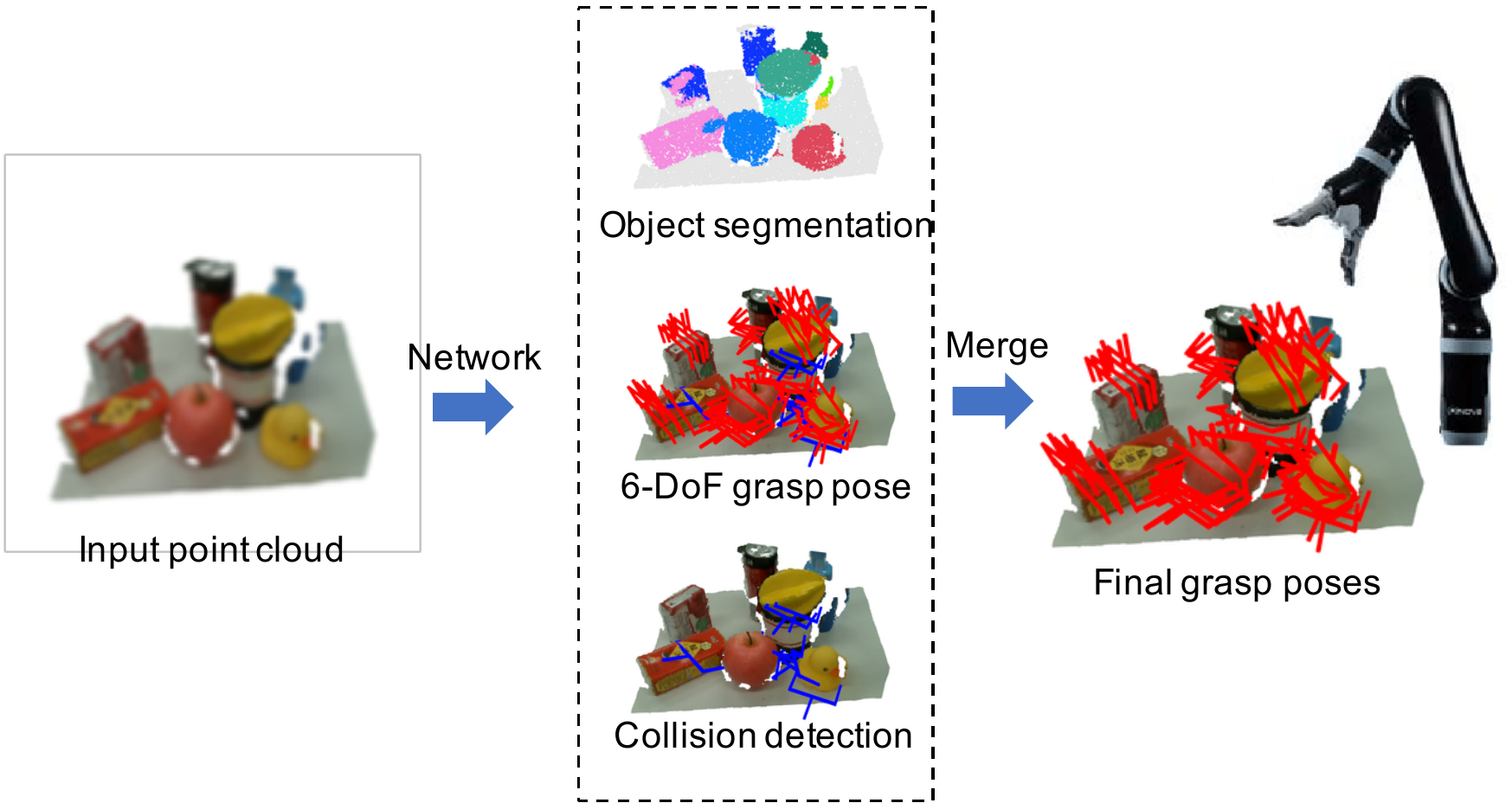}
    \caption{Overview of our proposed simultaneous multi-task learning framework. Given a scene point cloud, our network jointly predicts instance labels, 6-DoF grasp poses and collisions simultaneously. Finally, we merge three branches and the instance-level, collision-free 6-DoF grasp poses are generated for the robot to execute. }
    \label{fig:pipeline}
    \vspace{-2em}
\end{figure}


Recent works tend to directly regress 6-DoF grasp poses, or evaluate grasp quality scores from raw 3D point cloud data \cite{gualtieri2016high,ten2017grasp,liang2019pointnetgpd,mousavian20196,qin2020s4g,fang2020graspnet}.
They also utilize a collision detection module as post-processing to filter invalid grasps.
Despite their impressive results, 
we observe several potential drawbacks in such methods. 
a) They can not learn instance-level grasps. \textcolor{black}{The lack of instance information leads to the model can not carry out target-driven grasping, which is common in manipulation tasks. 
We also believe that the instance information could boost the grasp learning process.}
b) They always rely on 
collision detection as a post-processing module to filter invalid grasps, which is usually indispensable and time-consuming. 

\begin{figure*}[htbp]
    \centering
    \includegraphics[width=0.98\textwidth]{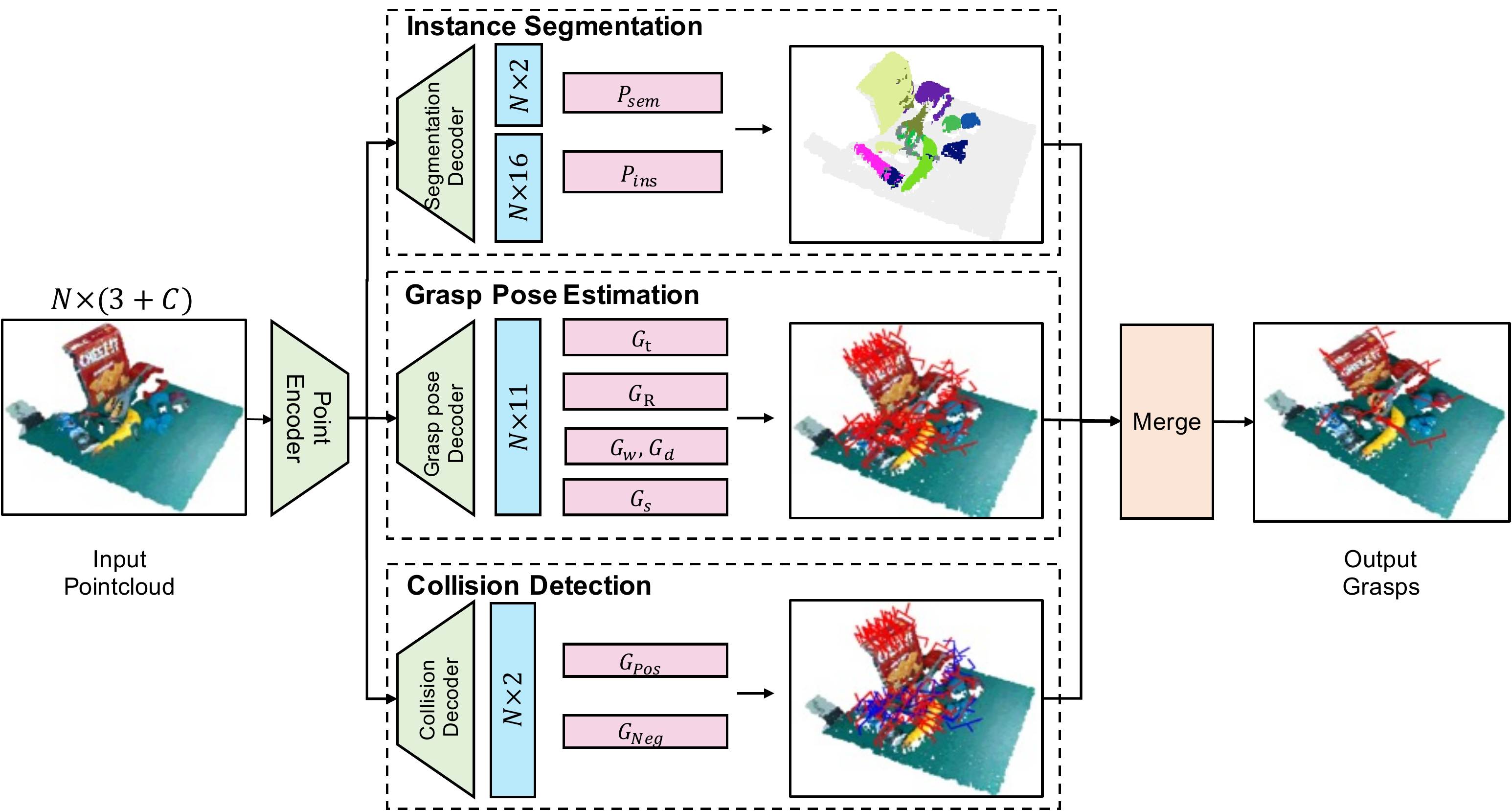}
    \caption{Framework of our proposed grasp pose learning method. Given N points with (3+C)-dim features, we adopt a point encoder to extract features and three decoders are used to predict object instances, grasp poses and collisions.$ P_{sem},P_{ins}$ denote the predicted  point-wise semantic, instance results, $G_t,G_R,G_w,G_d,G_s$ respectively mean center, rotation matrix, width, depth and score of predicted grasps, and $G_{pos},G_{neg}$ are predicted collisions defined by 6-DoF grasps. An instance based pose-NMS algorithm is used to merge these three modules to \textcolor{black}{form} final instance-level, collision-free grasps. } 
    \label{fig:framework}
    \vspace{-1em}
\end{figure*}

Ideally, a 6-DoF grasp pose estimation model should not only predict the feasible grasp poses but also be able to give the object level and collision information to guide the robot grasping.
In this paper, we propose a joint learning framework for 6-DoF grasping, which simultaneously predicts the instance semantic segmentation, feasible grasp poses, and potential collisions, as shown in Fig.~\ref{fig:pipeline}. Given a single-view point cloud as input, our model outputs object-level and collision-free 6-DoF grasps. 
More specifically, 
We adopt 3D PointNet++~\cite{qi2017pointnet++} as our backbone network to extract point features, then jointly train these three target branches in a unified manner, as shown in Fig.~\ref{fig:framework}. 
We observe that the position relationship between two points of a rigid object in 3D space is fixed, so the SE(3) grasps can be decomposed with two unit offsets, the approach direction and the close direction of a gripper.
The grasp poses are merged with the instance prediction and collision detection modules to form the final accurate, robust, and collision-free grasp poses.

We study the 6-DoF grasping problem
for a parallel gripper under a realistic yet challenging setting. 
Assuming various objects are scattered on a table, we only capture a single-view point cloud 
using a commodity camera. The camera pose is also unfixed so that a partial point cloud can be captured from a random viewpoint instead of the top-down view. This task is challenging both in perception and planning, caused by the scene clutters, multiple unknown objects, incomplete point cloud, and high grasp dimensions.

Extensive experiments on the public dataset~\cite{fang2020graspnet} and real-world robot systems demonstrate the effectiveness of our approach. 
The results show that both segmentation and collision branches boost the performance of grasp pose estimation. Semantic information improves grasp pose learning by identifying which instance a point belongs to and collision attributes help to filter invalid grasps. 
Our method also outperforms current start-of-the-art methods both on the dataset and real robot experiments.
In summary, our main contributions are as follows:
\begin{itemize}
    \item We propose a simultaneous multi-task learning paradigm for 6-DoF grasp pose estimation in structured clutter with instance semantic segmentation and collision detection.
    \item With the proposed simultaneous multi-task grasping, we could directly predict the point-wise 6-DoF grasp poses.
    \item Our proposed method outperforms state-of-the-art grasp pose estimation methods both on the large-scale dataset and real robot experiments.
\end{itemize} 

\section{Related Work}

In this section, we introduce the previous works related to ours.

\textbf{Learning based Grasping Methods.}
A large amount of prior grasp pose estimation methods focus on detecting a pair of graspable points or a rectangle on the planar surface based on RGB-D image input~\cite{lenz2015deep,depierre2018jacquard,chu2018real,jiang2011efficient,redmon2015real,levine2018learning,pinto2016supersizing,mahler2017dex2,guo2016object}. These simplified methods are constrained on a 2D plane since they only have 3-DoF
at most. 
In order to improve the dexterity of grasping, some researchers focus on predict 6-DoF grasp poses. 
A straightforward way is to apply 6-DoF object pose estimation to grasp tasks \cite{xiang2017posecnn,wang2019densefusion,deng2020self,wei2021gpr}. Although such model-based approaches can acquire accurate 6-DoF grasps, it requires prior knowledge about the geometry information of the object. 
Recently, some works directly 
predict 6-DoF grasp pose from point clouds. \textcolor{black}{GPD~\cite{ten2017grasp} proposes to generate grasp candidates based on the local geometry.
A CNN-based network is also proposed to classify generated grasps.
PointnetGPD~\cite{liang2019pointnetgpd} extends this work to 3D space and evaluates generated grasps via a PointNet-based network.}
S$^4$G~\cite{qin2020s4g} uses a single-shot grasp detection proposal network to regress $SE(3)$ grasps in densely cluttered scenes, and \cite{fang2020graspnet} designs a two-stage network to decouple grasp poses as approach vectors and other grasp operations (angle, width, depth, etc.). 
GraspNet~\cite{mousavian20196} adopts a variational autoencoder to sample grasps \textcolor{black}{for the single object} and introduces a grasp evaluate network to refine sampled grasps. 

In~\cite{murali20206}, the authors present a target-driven grasp method to grasp a specific object in structured clutters.
They use a cascaded pipeline to first crop the target object, then inference the grasp poses, and finally adopt a neural network to predict collision scores. The main difference between this work and ours is that we formulate the 6-DoF grasp task as a simultaneous multi-task learning problem, using a single shot grasp neural network, with jointly training segmentation and collision modules.

\textbf{Grasp Dataset Generation.}
Some previous works~\cite{jiang2011efficient,depierre2018jacquard} use rectangle representation for grasp detection annotated by humans. In~\cite{levine2018learning,pinto2016supersizing}, the authors collect grasp labels with real robot experiments.
Dexnet~\cite{mahler2017dex2} provides a universal grasp generation method by calculating force closure as grasp quality score from single object mesh.
Some followed studies~\cite{ni2020pointnet++,qin2020s4g} generate synthetic scene datasets by mapping 6-DoF object pose to single object grasps. 
\cite{eppner2019billion} compares a detailed grasp sampling strategies for data generation, and \cite{eppner2020acronym} collects a large-scale grasp dateset on simulation. 
To deal with the gap between the virtual environment and the real world, 
\cite{fang2020graspnet} constructs a general grasp dataset in cluttered scenes, in which images are captured by regular commodity cameras. 

\textbf{Deep Learning for Point Cloud.}
Some researchers analyze point clouds by projecting points to multi-view or volumetric representations~\cite{qi2016volumetric,wu20153d}. 
To preserve complete geometry information, PointNet~\cite{qi2017pointnet} and PointNet++~\cite{qi2017pointnet++} directly process raw points through employing multi-layer perceptron (MLP) and yield point-level features. Such point-based framework has been widely used in the 3D domain such as classification, segmentation and detection \cite{tatarchenko2017octree,graham20183d,shi2019pointrcnn}. 
Due to its inspiring performance, here we also adopt PointNet++ as our backbone network to extract features of the point cloud.

\section{Methods}


\textcolor{black}{Given a scene point cloud, we first use the backbone network PointNet++ \cite{qi2017pointnet++} to encode features, then simultaneously attach three parallel decoders: instance segmentation, 6-DoF grasp pose, and collision detection. These three heads respectively output predicted point-wise instances, grasp configurations, and collisions. At inference phase, grasps from the same instance without collision are grouped, and a pose non-maximum suppression algorithm is proposed to form the final grasps.}

\textcolor{black}{As shown in Fig.~\ref{fig:framework}, the input size of point cloud $\mathcal{P}$ is $N_p\times (3+C)$, where $N_p$ is the number of points, and $C=6$ is the extra channels about RGB colors and normalized  coordinates. The PointNet++ encoder module is composed of four set abstraction (SA) layers, while decoder layers are three feature propagation (FP) modules\footnote{For more details about the PointNet++, we refer readers to~\cite{qi2017pointnet++}.} following with MLP layers, represents the segmentation head, grasp head, and collision head. The output size of these three heads are $N_p\times (2+16)$ (semantic mask and instance embedding), $N_p\times (2+6+1+1+1)$ (graspable mask, two rotation vectors, grasp depth, width and score) and $N_p\times 2$ (collision mask). }




\subsection{Instance Segmentation Branch}\label{segmentataion}

We first attach a point-wise instance semantic segmentation module to distinguish multiple objects. 
Specifically, we formulate object instance segmentation as an embed-and-cluster task. 
Points belonging to the same instance should have similar features, 
while features for different instances should be dissimilar. 
\textcolor{black}{During training, the semantic and instance label of each point are known, and}
we supervise the outputs semantic labels with a two-class cross entropy loss $L_{sem}(s_i,\hat s_i)$ to classify background and foreground, where $s_i$ and $\hat s_i$ represent predicted and ground truth binary semantic labels. The instance loss is optimized through a discriminative loss function $L_{ins}$~\cite{de2017semantic}:

\begin{equation}\label{eq1}
\begin{aligned}
    L_{ins} &=  L_{var}+ L_{dist}+\alpha L_{reg}
\end{aligned}
\end{equation}

where 

\begin{equation}
\begin{aligned}
    L_{v a r}&=\frac{1}{C} \sum_{c=1}^{C} \frac{1}{N_{c}} \sum_{i=1}^{N_{c}}\left[\left\|x_{i}-\mu_{c}\right\|-\delta_{\mathrm{v}}\right]_{+}^{2}\\
    L_{\text {dist }}&=\frac{1}{C(C-1)} \sum_{c_{A}=1}^{C} \sum_{c_{B}=1}^{C}\left[2 \delta_{\mathrm{d}}-\left\|\mu_{c_{A}}-\mu_{c_{B}}\right\|\right]_{+}^{2}\\
    L_{r e g}&=\frac{1}{C} \sum_{c=1}^{C}\left\|\mu_{c}\right\|.
\end{aligned}
\end{equation}

Here $C$ is the number of objects, $N_c$ is the number of points in object $c$, \textcolor{black}{$x_i$ is the feature embedding of point $i$ belong to $c$, $\mu_c = \frac{1}{N_{c}} \sum_{i=1}^{N_{c}}x_{i}$ is the object feature center of class $c$, $||\cdot||$ means the L2 distance. $[x]_+ = max(0,x)$ denotes the hinge. $\delta_v$ and $\delta_d$ are two margins for variance and distance}
$L_{var},L_{dist},L_{reg}$ respectively represent variance loss, distance loss and regularization loss, \textcolor{black}{and $\alpha \ll 1$ is the regularization weights.} Variance loss means an intra-cluster loss that draws embeddings towards the cluster center, while distance loss is an inter-cluster loss that increases the distance between different cluster centers. Regularization loss constraints that all clusters towards the origin, to keep the activations bounded. 

The total segmentation loss $L_{seg}$ is defined as $L_{seg} = L_{sem} + L_{ins}.$ After learning the embeddings, a Meanshift \cite{comaniciu2002mean} clustering algorithm is applied to group points belong to the same instance. 

\subsection{6-DoF Grasp Pose Estimation Branch}\label{grasp}
\begin{figure}[htbp]
    \centering
    \includegraphics[width=0.4\textwidth]{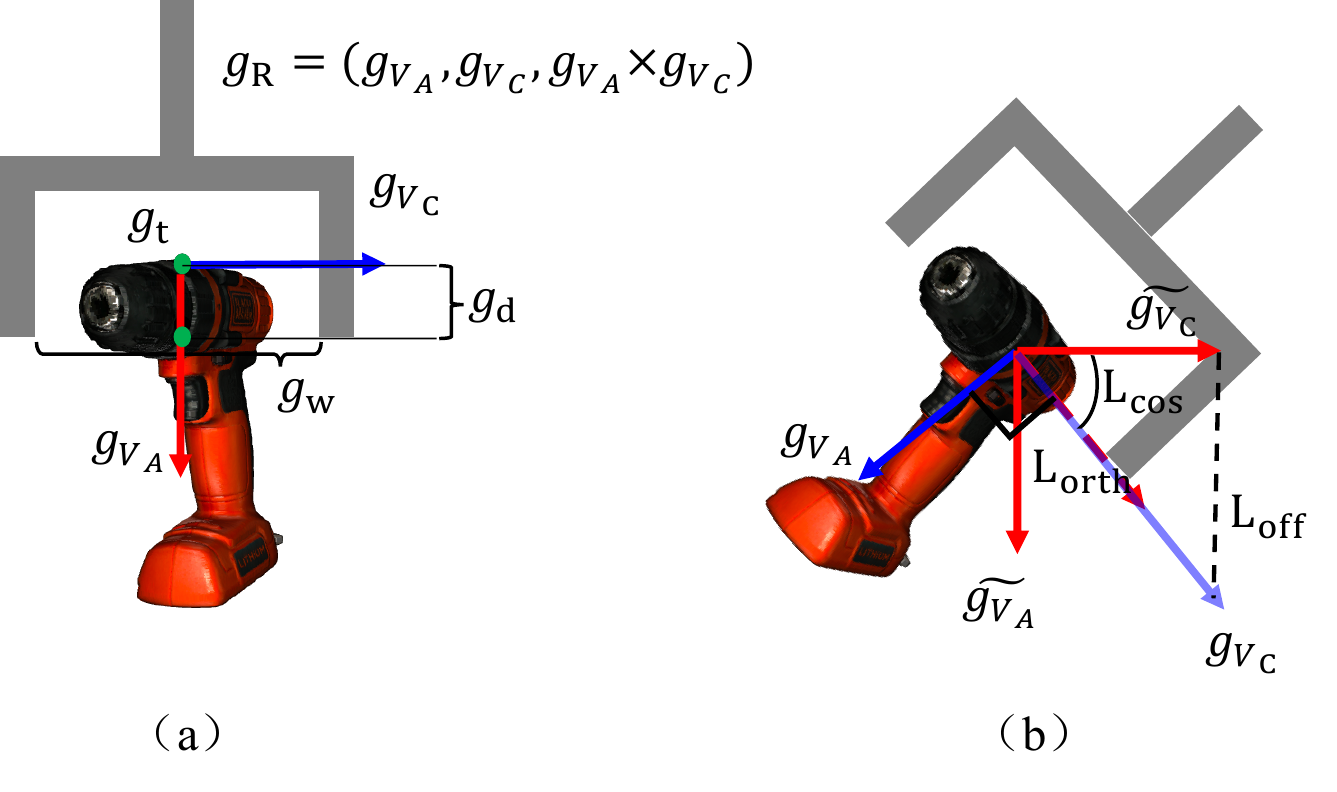}
    \caption{Grasp illustration. (a) Representation of 6-DoF grasp in $SE(3)$. (b) Representation of decomposed rotation loss $L_{rot}$.}
    \label{fig:grasp}
    \vspace{-1em}
\end{figure}
As shown in Fig.~\ref{fig:grasp} (a), the $SE(3)$ grasp configuration $g$ of the parallel gripper are formulated with a grasp center $g_t$, rotation matrix $g_R$, width $g_w$, depth $g_d$ and a grasp score $g_s$:
\begin{equation}
    \textcolor{black}{g=(g_t,g_R,g_w,g_d,g_s),}
\end{equation}
where $g_s$ is used to evaluate the quality of a grasp computed by a improved force-closure metric \cite{liang2019pointnetgpd}.
To directly regress the point-wise 6-DoF grasp pose, we introduce two assumptions for grasp training:
\begin{itemize}
    \item For each point $p$ in the point cloud, there is at most one good grasp $g$, such that the mapping $f:p \rightarrow g$ is unique and fixed.
    \item The grasp center $g_t$ is defined on the object surface and refined by grasp depth $g_d$. That means it just needs to classify graspable points to predict grasp centers, instead of regressing 3D coordinates. 
\end{itemize}

Directly learning the 6-DoF grasp pose through rotation representation such as quaternions or rotation matrices is challenging since they are nonlinear and discontinuous.
To handle this problem, we propose a vector based 
method that decompose the rotation matrix with two orthogonal unit vectors.For a grasp $g$, The rotation matrix \textcolor{black}{$g_R \in \mathcal{R}^ {3 \times 3}$} is defined as:
\begin{equation}\label{eq6}
    g_R = [g_{v_A}, g_{v_C}, g_{v_A} \times g_{v_C}],
\end{equation} 
\textcolor{black}{where $g_{v_*} \in \mathcal{R}^{3 \times 1}$ is a column vector.} $g_{v_A}$ limits the approach direction of gripper, and $g_{v_C}$ is the direction of gripper closing. 

For optimization, we divide the rotation loss $L_{rot}$ into three parts (Fig.~\ref{fig:grasp}(b)): offset loss $L_{off}$, cosine loss $L_{cos}$, and a relate loss $L_{orth}$, which respectively constrains the position, angle prediction, and the orthogonality:
\begin{equation}\label{eq7}
    L_{rot}(g_{v_*}, \hat g_{v_*}) = \beta_1 \cdot L_{off} +\beta_2 \cdot L_{cos} +\beta_3 \cdot L_{orth},
\end{equation}

\begin{equation}
\begin{aligned}
    L_{off}(g_{v_*}, \hat g_{v_*}) &= \frac{1}{G} \sum_{g \in G} \left\|g_{v_*} - \hat g_{v_*}\right\|\\
    L_{cos}(g_{v_*}, \hat g_{v_*}) &= -\frac{1}{G} \sum_{g \in G} |g_{v_*} \cdot  \hat g_{v_*}|\\
    L_{orth}(g_{v_A}, g_{v_C}) &= -\frac{1}{G} \sum_{g \in G} |g_{v_A} \cdot g_{v_C}|,
\end{aligned}
\end{equation}
where $g$ is a predicted grasp in set $G$, $g_{v_*}, \hat g_{v_*}$ are ground truth and predicted vectors. \textcolor{black}{$\beta_1,\beta_2,\beta_3$ are three coefficients to balance different terms.} As the gripper close direction is symmetric, we limit $g_{v_C} = -g_{v_C}   \indent \text{if} \indent g_{v_C}\cdot (1,0,0)^T <0$. We observe that this decomposed loss representation can improve the accuracy of grasp rotation compared with directly calculate the distance between two quaternions.

The prediction of graspable point is treated as a two-class classification task and a cross-entropy loss function \textcolor{black}{$L_{gp}(g_{p_i},\hat g_{p_i})$} is applied \textcolor{black}{with weight $w_{1}=1.0$ and $w_{2}=5.0$. $g_{p_i}, \hat g_{p_i}$ donate the binary ground truth and prediction of graspable or not for point $i$. In addition, we regress grasp width, depth and score for each grasp $g$ with mean squared error (MSE) loss, record as $L_w(g_w,\hat g_w)$, $L_d(g_d,\hat g_d)$ and $L_s(g_s,\hat g_s)$. $g_*$ and $\hat g_*$ respectively mean grasp labels and predicted grasps.} The total grasp loss is written as:

\begin{equation}\label{eq11}
\begin{aligned}
    L_{grasp} =\sum_{P_s} L_{gp} +  \sum_{P_g} (L_{rot} + \gamma_1 L_{w} + \gamma_2 L_{d} + \gamma_3 L_{s}),
\end{aligned}
\end{equation}
where $P_s$, $P_g$ represent the whole scene point cloud and point set with graspable points in ground truth, respectively. \textcolor{black}{$\gamma_1,\gamma_2,\gamma_3$ are coefficients for grasp configurations}.

\subsection{Collision Detection Branch}\label{collision}

Although the methods above are able to predict 6-DoF grasps at the instance level, it is still necessary to determine whether generated grasps are valid and executable in scenes.
\textcolor{black}{For this purpose, we attach a collision detection branch to infer potential collisions for each grasp.}

We provide a learn-able collision detection network to directly
predict collisions for generated grasps. \textcolor{black}{Binary} collision labels are generated by an off-the-shelf collision detection module according to the grasp configuration. \textcolor{black}{During training, we sample both positive (without collisions)  and negative (with collisions) grasps and supervise the collision as a classification task with a two-class cross-entropy loss $L_{coll}(c_i,\hat c_i)$:}
\begin{equation}\label{eq12}
    \textcolor{black}{L_{coll} = \sum_{P_i} -[c_i\cdot \log \hat c_i + (1-c_i)\cdot \log (1-\hat c_i)],}
\end{equation}
\textcolor{black}{where $c_i$ is the collision label and $\hat c_i$  denotes the predicted collision probability for point $i$.}


\begin{algorithm}\label{nms}
\textbf{Input:} Prediction $P = {(S,G,C)}$\\
\quad S, G, C are sets of predicted semantic, grasp pose and collision labels, respectively.\\
\textbf{Export:} Executable grasps set $V=\{ \}$.

Select collision-free grasps $G_C = G \cap \tilde C$.

\For{$(G_i \subset G_C) \ where \ (S_i \subset instance\ N)$}{
\While{ $G_i != \emptyset$}{
\textcolor{black}{
Sort $g \in G_i \ by \ g_{s}$\\
$g = g_0$ \\
Add $g\ to\ V$, Delete $g_0$ \\
\For{$g_k \in G_i$ )}{
\If{$SE(3)  Distance(g,g_k) < \epsilon$}{
    Delete $g_k$ \\}
}
}
}
}
\textbf{Output:} Executable grasps set $V = \{g_1,g_2,\dots g_{n}\}$
\caption{Instance based Pose-NMS}
\end{algorithm}
\subsection{Forming Final Grasps}

For training, the total loss of the whole multi-task learning framework is written as:
\begin{equation}\label{eq13}
    L = L_{seg} + L_{grasp} + L_{coll}.
\end{equation}

After three branches being jointly optimized, \textcolor{black}{our network outputs the instance label, grasp, and collision for each point. Then we run the instance-based pose-NMS algorithm to merge three modules and form final executable grasps, as shown in Algorithm~\ref{nms}.} Points belong to the same instance are grouped and we calculate the SE(3) distance between two grasps to suppress non-maximum grasps. \textcolor{black}{The distance threshold $\epsilon$ is set to $10mm, 30\degree$.}


\section{Experiments}

In this section, 
we first introduce the experimental setup, including dataset, metrics, and implementation details. 
Then we analyze the main results and perform ablation studies to evaluate the effectiveness of different modules. Finally, robot experiments are conducted to validate the performance of our method in real-world robot grasping.

\subsection{Experimental Setup}

\subsubsection{Dataset and Metrics}\label{metrics}
We evaluate the proposed method on both public benchmarks and real robot systems.
We adopt GraspNet-1Billion dataset for benchmark evaluation~\cite{fang2020graspnet}, which is a large benchmark for general object grasping. The dataset contains 97,280 RGB-D images with over one billion 6-DoF grasp poses for 88 objects, captured by two popular cameras, Intel Realsense and Kinect in the real world. 
It is split into the train, test seen, test similar, and test novel sets with 100, 30, 30, 30 scenes, and each scene contains 256 images and millions of grasps for 10 objects randomly sampled. 
We adopt average \textit{Precision@k} as the evaluation metric \cite{fang2020graspnet} , which measures the precision of top-$k$ ranked grasps. 
$AP_\mu$ denotes the average precision for $k$ ranges with a force closure parameter lower than $\mu$, where $\mu \in \{ 0.2,0.4,0.6,0.8,1.0,1.2 \}$. Here we use the default value of parameter $k=50$.
In real robot experiments, we adopt grasp success rate and completion rate to evaluate the performance. Success rate denotes the percentage of successful grasps, and completion rate represents the percentage of objects that is successfully grasped. 

\subsubsection{Implementation Details}
The input point cloud is generated from an RGB-D image, and preprocessed with workspace filtering, random sampling, and normalization. The number of points $N_p$ is set to 20,000. 
Because a large amount of grasps for each image is too dense for the network to learn, 
we select object grasps with $score > 0.5$ by approach-based grasp sampling~\cite{eppner2019billion} during training.
The point-grasp mapping is formulated by calculating the minimum distance between points and grasp centers within $5 mm$.
Points without any ground-truth grasps are ignored, which contributes nothing to the training objective. 
Our neural network is trained for 80 epochs with Adam optimizer~\cite{kingma2014adam}. The initial learning rate is 0.05 and decreases by a factor of 2 for every 10 epochs \textcolor{black}{with the batchsize of 64.} \textcolor{black}{For the loss function, we set $\alpha=0.01$, $\delta_d,\delta_v=1.5,0.5$, $\beta_1,\beta_2,\beta_3 = 5.0,1.0,1.0$, and $\gamma_1,\gamma_2,\gamma_3 = 100,1000,10$.} 

\textcolor{black}{At inference, our network outputs point-wise semantic, instance embeddings, grasps, and collisions. A MeanShift algorithm is adopted to group points belong to the same instance.s
For each predicted graspable point, a grasp configuration composed of grasp pose, width, depth, and score is generated. Collision detection network outputs the probability of whether a grasp is collision-free.}

\textcolor{black}{
The proposed simultaneous 6-DoF grasp pose estimation model consists of a point encoder module and three decoders: segmentation head, grasp head and collision head. We describe the network architecture of our model and summarize the details as Tab.~\ref{tab:netarch}. Following the same notation in PointNet++, $SA(K,r,[l_1,\cdots,l_d])$ is a set abstract layer with $K$ local regions in radius $r$, and [$l_1,\cdots,l_d$] are fully connected layers with $l_i(i=1,\cdots,d)$ output channels. FP([$l_1,\cdots,l_d]$) is a feature propagation layer with $d$ fully connected layers. 
MLP($l_1,\cdots,l_d$) means a multi-layer perceptron with output layer sizes $l_1,\cdots,l_d$ on each point.}

\begin{table}[h]
\color{black}
\caption{Detailed network architecture of our proposed method.}
\label{tab:netarch}
\begin{center}
\begin{tabular}{cc}
\hline
Network & Architecture \\
\hline
Point encoder & \makecell{SA(1024,0.1,[32,32,64]) $\rightarrow$ \\ SA(256,0.2,[64,64,128]) $\rightarrow$ \\ SA(64,0.4,[128,128,256]) $\rightarrow$ \\ SA(None,None,[256,256,512])}\\
\hline
Segmentation head & \makecell{FP[256,256]$\rightarrow$FP[256,256]$\rightarrow$ \\
FP[256,128]$\rightarrow$FP[128,128,128]$\rightarrow$ \\
MLP(128,128,2) \& \\ MLP(128,128,16)} \\
\hline
Grasp head & \makecell{FP[256,256]$\rightarrow$FP[256,256]$\rightarrow$ \\
FP[256,128]$\rightarrow$FP[128,128,128]$\rightarrow$ \\
MLP(128,128,2) \& \\ MLP(128,128,6)\& \\ MLP(128,128,3)}
\\
\hline
Collision head & \makecell{FP[256,256]$\rightarrow$FP[256,256]$\rightarrow$ \\
FP[256,128]$\rightarrow$FP[128,128,128]$\rightarrow$ \\
MLP(128,128,2)} \\
\hline

\end{tabular}
\end{center}
\end{table}

\subsection{Main Results}





\begin{table*}[t]
\caption{6 Dof grasp pose estimation results on GraspNet-1Billion dataset.}
    \vspace{-1em}
\label{main_results}
\begin{center}
\begin{tabular}{c|c|c c c|c c c|ccc}
\hline
\multirow{2}*{Methods} &\multirow{2}*{\textcolor{black}{Grasp type}} &\multicolumn{3}{c}{Seen}&\multicolumn{3}{c}{Similar}&\multicolumn{3}{c}{Novel}\\
\cline{3-11}
{~} & {~} & \textbf{AP} & AP$_{0.8}$ & AP$_{0.4}$& \textbf{AP} & AP$_{0.8}$ & AP$_{0.4}$& \textbf{AP} & AP$_{0.8}$ & AP$_{0.4}$\\
\hline
GG-CNN \cite{morrison2018closing} & \textcolor{black}{2D planar} & 15.48 & 21.84 & 10.25& 13.26 & 18.37 & 4.62& 5.52 & 5.93 & 1.86\\

Chu et al.~\cite{chu2018real} & \textcolor{black}{2D planar} & 15.97 & 23.66 & 10.80& 15.41 & 20.21 &7.06 & 7.64 & 8.69 & 2.52\\

GPD \cite{ten2017grasp} & \textcolor{black}{6-DoF} & 22.87 & 28.53 & 12.84& 21.33 & 27.83 &9.64 & 8.24 & 8.89 &2.67\\

PointnetGPD \cite{liang2019pointnetgpd} & \textcolor{black}{6-DoF} & 25.96 & 33.01 & 15.37& 22.68 & 29.15 &10.76 & 9.23 &9.89 &2.74\\

GraspNet \cite{fang2020graspnet}  & \textcolor{black}{6-DoF} & 27.56 & 33.43 & 16.95& 26.11 & 34.18 &14.23 & 10.55 &11.25 &3.98\\

GraspNet re-implementation \cite{fang2020graspnet} & \textcolor{black}{6-DoF} & 32.47 & 37.72 & \textbf{29.21}& 26.78 & 31.96 &\textbf{23.19} & 10.27 &12.94 &\textbf{5.41}\\
\hline
Ours  & \textcolor{black}{6-DoF} & \textbf{36.55} & \textbf{47.22} & 19.24 & \textbf{28.36} & \textbf{36.11} &10.85 & \textbf{14.01} &\textbf{16.56} & 4.82\\

\hline
\end{tabular}
\end{center}
\end{table*}

\begin{table*}[t]
\caption{Ablation study on instance segmentation and collision detection modules6.}
    \vspace{-1em}
\label{tab:ablation}
\begin{center}
\begin{tabular}{l|c c c|c c c|c c c}
\hline
\multirow{2}*{Branches} &
\multicolumn{3}{c}{Seen}&\multicolumn{3}{c}{Similar}&\multicolumn{3}{c}{Novel}\\
\cline{2-10}
{~} & \textbf{AP} & AP$_{0.8}$ & AP$_{0.4}$& \textbf{AP} & AP$_{0.8}$ & AP$_{0.4}$& \textbf{AP} & AP$_{0.8}$ & AP$_{0.4}$\\
\hline
Pose & 25.57 &31.46 & 13.75& 16.01 & 19.52 & 6.97 & 7.55 & 8.34& 2.40\\

Pose+Seg & 26.62 & 33.94 & 14.72& 16.93 & 21.27 &7.64 & 8.43 & 9.68 & 2.93\\

Pose+Coll & 34.58 & 43.96 & 18.42& 27.35 & 33.86 &\textbf{12.30} & 12.07 & 14.08 & 4.00\\

Pose+Seg+Coll & \textbf{36.55} & \textbf{47.22} & \textbf{19.24}& \textbf{28.36} & \textbf{36.11} &10.85 & \textbf{14.01} & \textbf{16.56} & \textbf{4.82}\\
\hline
\end{tabular}
\end{center}
\end{table*}

\begin{figure*}[t]
    \centering
    \includegraphics[width=0.8\textwidth]{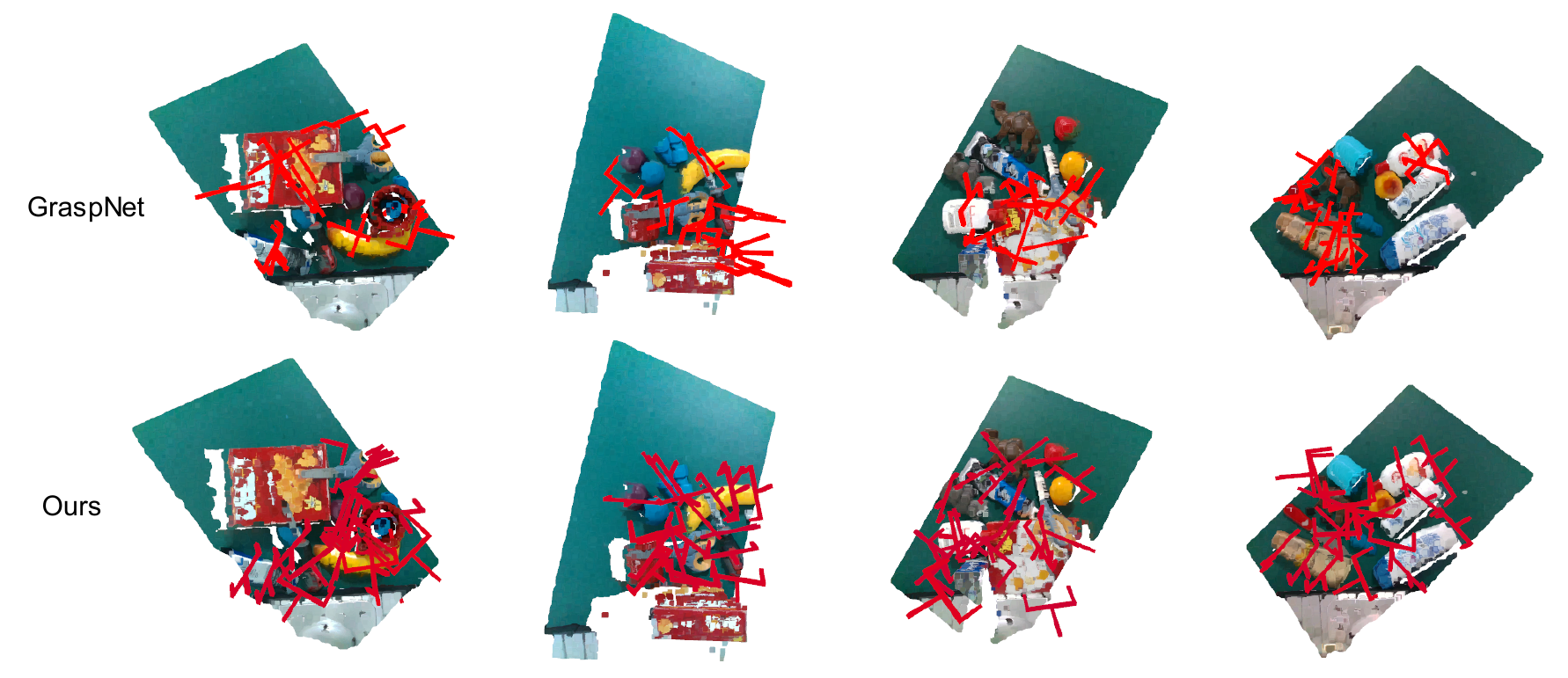}
    \caption{A visual comparison for GraspNet~\cite{fang2020graspnet} and our proposed grasp method. It can be seen that our model could generate high quality, more diverse grasp poses on object level.}
    \label{fig:compare_graspnet}
        \vspace{-1em}
\end{figure*}

We compare the performance 
of different methods
on GraspNet-1Billion, which is based on physical analysis with calculating the grasp quality by force closure metric (Tab. \ref{main_results}). 
It can be seen that our simultaneous learning approach achieves the best performance. \textcolor{black}{Specifically, compared with rectangle-based grasp methods which execute top-down grasps on the 2D plane~\cite{morrison2018closing,chu2018real}, the 6-DoF grasp representation has obvious advantages. Besides, point-based networks also outperform 2D convolution networks~\cite{ten2017grasp}.} 
Compared with other deep 6-DoF grasp methods~\cite{liang2019pointnetgpd,fang2020graspnet}, our approach also shows obvious gains (\eg~+4.08 AP on seen split).

Despite our single-shot end-to-end network predicts grasps without any refinement, \textcolor{black}{it still has the best performance of mAP on all of the test datasets.}
In addition, we observe that our model always performs better on $AP_\mu$ with a large $\mu$, indicating that 
it tends to generate more proper predictions and avoids wrong candidates to ensure an executable grasp instead of learning several best grasps only, which is significant in real robot grasping tasks.

For qualitative valuation, we visualize the outputs of grasp pose results and compare them with the benchmark method. For a fair comparison, we choose 20 grasps with the highest scores in each point cloud, which is shown in Fig. \ref{fig:compare_graspnet}. 

\subsection{Ablation Study}

Our multi-task learning method introduces instance segmentation and collision detection modules to improve grasp pose estimation. 
To investigate the influence of these components and the proposed rotation loss, 
we conduct a set of ablation studies.

\textit{Instance Segmentation.}
The instance segmentation information boosts the naive grasp pose baseline by 1.05\%, as shown in Tab.~\ref{tab:ablation}. 
The improved performance demonstrates that the segmentation module boosts grasp pose learning by providing more semantic information about instance geometry and relation. 
Note that different objects are usually occluded with each other, the segmentation plays a crucial role in separating different objects in the cluttered scenario.

\textit{Collision Detection.}
Another important feature of our framework is that it could simultaneously predict the collision probability of each potential grasp pose. The collision module prevents grasps with invalid poses which are misjudged by the network. Including collision detection branch improves the pose AP by 9.01\%, which demonstrates its effectiveness. 

\textit{Rotation Loss.}
Our proposed $L_{rot}$ loss also benefits grasp pose learning by 1.59\%, as shown in Tab. \ref{rotationloss}. \textcolor{black}{We compare it with directly calculating the related angle between quaternion label $g_{q}$ and prediction $\hat g_{q}$, where the loss function $L_{quat}$ is defined by:}
\begin{equation}\label{eq14}
\begin{aligned}
    Angle(g_q, \hat g_q) &= \arccos (0.5 \times (\mathrm{trace}~[g_{R}  {\hat g_{R}}^T] -1) \\
    L_{quat} &= \frac{1}{G} \sum_{g \in G} Angle(g_q, \hat g_q),
\end{aligned}
\end{equation}
\textcolor{black}{where $g_{R}, \hat g_{R}$ are two rotation matrices equal to $g_{q}, \hat g_{q}$.}
By decomposing rotation matrix with two specific directions, our network learns the nonlinear representation much easier. 

\begin{table}[h]
\caption{Ablation study on rotation loss.}
\label{rotationloss}
\begin{center}
\begin{tabular}{c|c|c|c}
\hline
Loss & AP & AP$_{0.8}$ & AP$_{0.4}$ \\
\hline
\textcolor{black}{Quaternion angle} & 34.96 & 45.39 & 18.53\\
\hline
Ours & \textbf{36.55} & \textbf{47.22} & \textbf{19.24}\\
\hline
\end{tabular}
\end{center}
\end{table}

\textit{Running Time.} The inference time for the forward path of our network on Nvidia RTX 2070 is 21ms, which could largely satisfy the most application requirements. \textcolor{black}{The post-processing time, including point MeanShift clustering and grasp pose-NMS, respectively cost 762ms and 851ms on Intel Core i5-8500 CPU. 
The total time efficiency of our method is 1634ms. 
We believe both clustering and pose-NMS could be implemented in more time-efficient ways~\cite{jiang2020pointgroup,wang2020solov2}, which will be further explored in the future work.
}

\subsection{Robot Experiment}

\begin{figure}[htbp]
    \centering
    \includegraphics[width=0.3\textwidth]{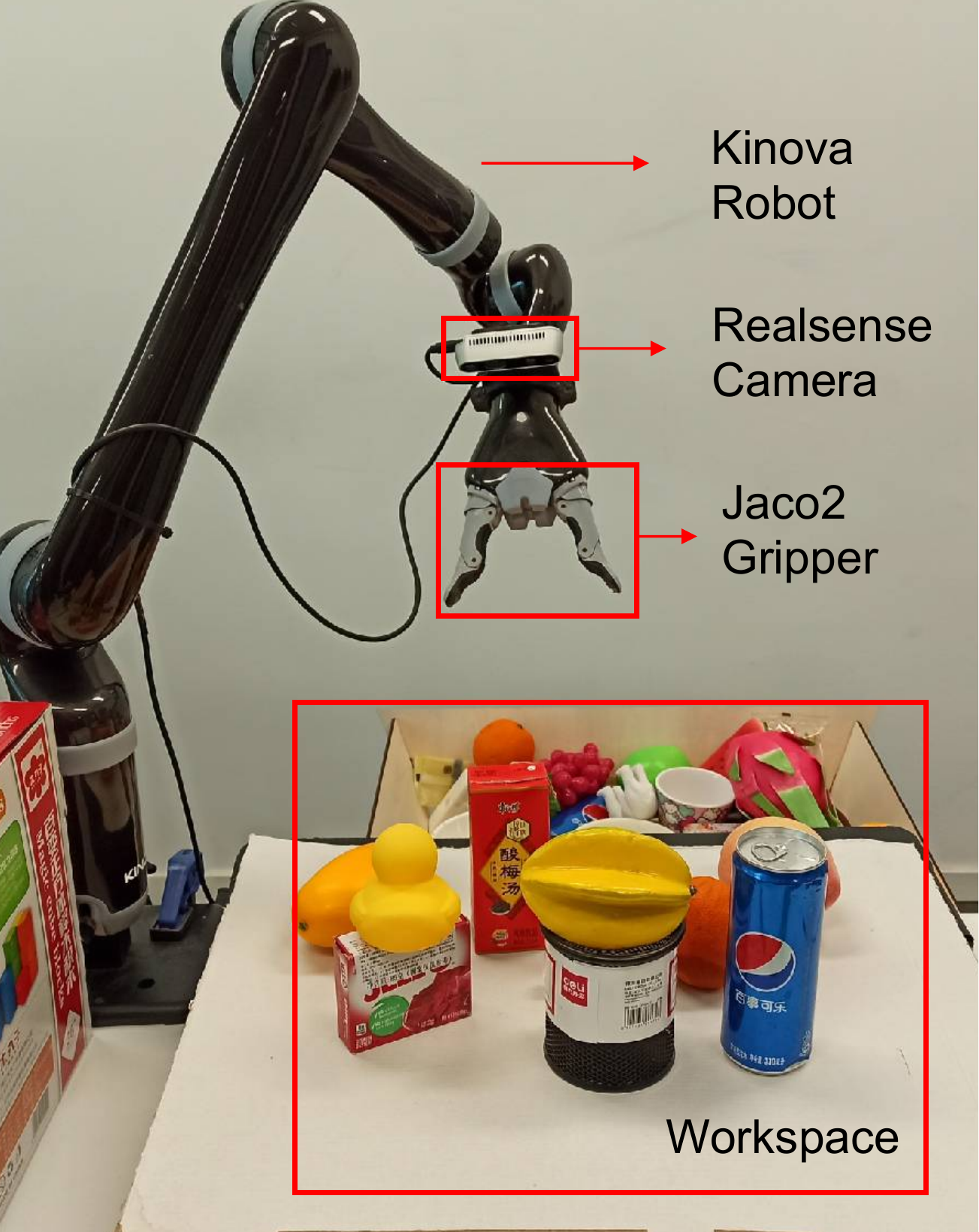}
    \caption{Our robot grasp system and workspace.}
    \label{fig:workspace}
        \vspace{-1em}
\end{figure}

In order to evaluate the performance of our proposed methods in the real world, we set up robot experiments on a Kinova Gen2 Robot with a parallel gripper Jaco2. 
A commercial RGB-D camera, Realsense D435i is mounted on the robot wrist to capture the input point cloud from an oblique perspective (Fig.~\ref{fig:workspace}).

We prepare over 20 objects absent in the training dataset with various shapes for robot grasping. For each experiment procedure, we randomly choose 5, 8, 10 objects, and put them on a table to form a structure cluttered scene. The point cloud of the scene is firstly collected by the Realsense camera and sent to the trained model to get the final grasp parameters.
The robot attempts multiple grasps until all of the objects are grasped, and the max number of grasping attempts is set to 15. We repeat the experiments 3 times for each method. 
We calculate the success rate in Tab.~\ref{robotexp}. \textcolor{black}{It can be seen that compared with 2D planar grasp (GQ-CNN~\cite{mahler2017dex2}), 6-DoF grasp from diverse directions performs better.}
 Our model shows a 76.0\% success rate with an 82.6\% completion rate and outperforms baseline methods by a large margin. 
 It proves that our grasp prediction model can successfully transfer to real robot grasp tasks.
 
 \begin{table}[t]
\caption{The main results of real robot experiment.}
        \vspace{-1em}
\label{robotexp}
\begin{center}
\begin{tabular}{c|c|c}
\hline{Methods} &Success rate & Completion rate \\
\hline
\textcolor{black}{GQ-CNN~\cite{mahler2017dex2}}  & \textcolor{black}{58.1\%} & \textcolor{black}{62.3\%} \\
\hline
GPD (3-channel)~\cite{ten2017grasp} & 61.5\% & 69.6\% \\
\hline

GPD (12-channel)~\cite{ten2017grasp} & 60.0\% & 65.2\% \\
\hline
\textcolor{black}{GraspNet~\cite{fang2020graspnet}}  & \textcolor{black}{72.4\%} & \textcolor{black}{79.7\%} \\
\hline
Ours& \textbf{76.0\%} & \textbf{82.6}\%\\
\hline
\end{tabular}
\end{center}
        \vspace{-1em}
\end{table}
 
\textcolor{black}{Since 2D planar grasp methods capture vision inputs from a top-down view, we also compare our proposed model with GQ-CNN~\cite{mahler2017dex2} from the same view for a fair comparison. Experiment results are recorded in Tab.~\ref{robotexptd}. Although the drastic gradient change in depth image boosts the 2D planar grasping, our simultaneous grasp learning method still has a better performance. We also observe that the top-down view brings severe
self-occlusion which leads to a worse point cloud input and causes more collision between grippers and the unseen part of the object, reducing the success rate and completion rate of grasping. It also proves that capturing visual inputs from an oblique viewpoint could improve the performance of 6-DoF grasping.}

\begin{table}[t]
\color{black}
\caption{Top-down view grasping experiment results.}
        \vspace{-1em}
\label{robotexptd}
\begin{center}
\begin{tabular}{c|c|c}
\hline{Methods} &Success rate & Completion rate \\
\hline
GQ-CNN~\cite{mahler2017dex2} & 64.1\% & 72.5\% \\
\hline
Ours&  \textbf{70.1\%} & \textbf{78.3\%} \\    
\hline
\end{tabular}
\end{center}
        \vspace{-1em}
\end{table}

\section{Conclusion}
In this paper, 
we formalize the 6-DoF grasp pose estimation in clutters as a simultaneous multi-task learning problem. 
We jointly predict the feasible 6-DoF grasp poses, instance semantic segmentation, and collisions. 
The whole framework is end-to-end trainable and jointly optimized in a unified network.
On the large scale public dataset, 
our method outperforms prior state-of-the-art methods by a large margin.
We also demonstrate the implementation of our model on a real robotic platform. \textcolor{black}{Besides, it is convenient to extend our simultaneous grasp learning model to several target-driven robotic manipulation tasks, such as picking and placement, object rearrangement, and interactive grasping. In the future work, we will (a) improve the pose-NMS efficiency by parallel implementations, and (b) utilize multi-view reconstruction techniques to further boost the performance.}

\addtolength{\textheight}{-1cm}   


\bibliographystyle{IEEEtran}
\bibliography{IEEEabrv,reference}

\end{document}